\documentclass{INTERSPEECH2024}
\usepackage{multirow}
\usepackage{stfloats} 
\usepackage{xcolor}
\usepackage{amssymb}
\usepackage{pifont}
\newcommand{\cmark}{\ding{51}}%
\newcommand{\xmark}{\ding{55}}%
\usepackage{caption}

\usepackage{varwidth}
\DeclareCaptionFormat{myformat}{%
  \begin{varwidth}{\linewidth}%
    \centering
    #1#2#3%
  \end{varwidth}%
}

\interspeechcameraready

\title{COSMIC: Data Efficient Instruction-tuning For Speech In-Context Learning}
\name[affiliation={1}]{Jing}{Pan}
\name[affiliation={1}]{Jian}{Wu}
\name[affiliation={1}]{Yashesh}{Gaur}
\name[affiliation={1}]{Sunit}{Sivasankaran}
\name[affiliation={1}]{Zhuo}{Chen}
\name[affiliation={1}]{Shujie}{Liu}
\name[affiliation={1}]{Jinyu}{Li}
\address{$^1$Microsoft, One Microsoft Way, Redmond, USA}
\email{\{jingpan,wujian,yashesh.gaur,sunit.sivasankaran,zhuc,shujliu,jinyli\}@microsoft.com}

\begin{document}

\maketitle
 
\begin{abstract}

We present a cost-effective method to integrate speech into a large language model (LLM), resulting in a Contextual Speech Model with Instruction-following/in-context-learning Capabilities (COSMIC) multi-modal LLM. Using GPT-3.5, we generate Speech Comprehension Test Question-Answer (SQA) pairs from speech transcriptions for supervised instruction tuning. With under 30 million trainable parameters and only 450 hours of English speech data, COSMIC demonstrates emerging capabilities in instruction-following and in-context learning. Equipped with such capabilities, COSMIC achieves a maximum 33.18 BLEU score in 0-shot EN-to-X speech to text translation (S2TT) and a significant boost in the 1-shot setting. Additionally, there is an average 25.8\% relative Word Error Rate (WER) reduction for 1-shot cross-domain adaptation. COSMIC exhibits a significant automatic speech recognition (ASR) accuracy gain in contextual biasing tasks due to its instruction-following capability.


\end{abstract}


\section{Introduction}

Recent developments in the field of natural language processing (NLP) have witnessed a surge in interest surrounding large language models (LLMs) \cite{brown2020language,touvron2023llama,anil2023palm} capable of contextualized learning. These models are often pre-trained on extensive corpora and exhibit remarkable proficiency in capturing intricate semantic relationships within language. The effectiveness of LLMs can be further augmented through in-context learning, a paradigm that focuses on adapting models to specific contextual nuances \cite{dong2022survey}. 

In the realm of speech processing, the acquisition of in-context learning capabilities has been a longstanding aspiration for researchers in the community \cite{wang2023can,hsu2023exploration}. Models dedicated to speech processing tasks, such as automatic speech recognition (ASR) and speech-to-text translation (S2TT), predominantly adhere to the supervised training paradigm \cite{rubenstein2023audiopalm,barrault2023seamlessm4t}. As a result, in order to train a speech model to accomplish a specific task, it is mandated to prepare a certain amount of paired speech-text data, which is not cost-effective. Even with self-supervision learning \cite{baevski2020wav2vec, hsu2021hubert}, the pre-trained models do not have any in-context learning capabilities as they still have to be further fine-tuned using supervised data to make it work for any particular tasks \cite{yang21c_interspeech}. We envision opportunities in advancing speech in-context learning by incorporating speech modality with the pre-trained foundation text LLMs, as these LLMs already exhibit in-context learning capabilities. Although some of the previous works \cite{fathullah2023prompting,lakomkin2023end,wang2023slm,yu2023connecting,wu2023decoder} validate the possibility of integrating the speech modality with LLMs, there have only been a limited number of attempts for in-context learning. Furthermore, an extensive evaluation of how the in-context abilities generalize to the speech domain has been missing. Inspired by instruction-following fine-tuning for visual-language models \cite{liu2023visual}, we aim to develop speech in-context learning capability through minimal instruction tuning data.

To prepare the instruction tuning with the speech data, we prompt GPT-3.5 to generate comprehension test question-answer (SQA) pairs based on the transcripts of our training corpus and train a speech encoder and a text LLM together to answer questions based on the input speech content. By instruction-tuning on 450 hours of English (EN) only speech data, COSMIC evolves the ability of zero-shot text-instruction following capability in ASR and zero-shot \& few-shot in-context learning in unseen EN$\to$X speech translation task. Our key contribution of this work lies in: a) \textbf{A data-efficient instruction-tuning method} which effectively glues speech input into pre-trained text LLMs and brings in-context learning abilities for unseen speech tasks. b) Thorough investigation and \textbf{quantitative analysis of the in-context learning} in speech tasks of the model.

We discuss about the related work in Section \ref{section:relatedwork} and present our methods in Section \ref{section:methodolgy}. We describe in details how the instruction-following fine-tune data is generated and the speech tasks we evaluate on in Section \ref{section:datasetup}. We dive deep into experiments and results in Section \ref{section:experiment}. Section \ref{section:conclusion} concludes the paper.

\section{Related work}
\label{section:relatedwork}


There has been growing interest in building a LLM incorporated with speech modality. SpeechGPT \cite{zhang2023speechgpt} maps speech signal to discrete units via a HuBERT-based \cite{hsu2021hubert} encoder and produce speech output through a vocoder, there by showcasing cross-modal instruction-following of the large language model. One drawback of the work is that the model's performance is not evaluated in a quantifiable metrics. AudioPaLM \cite{rubenstein2023audiopalm} followed a similar idea but also infused various speech tasks into a LLM, thereby benefiting from its strong language modeling capabilities. However, AudioPaLM did not show any signs of strong in-context learning  or generalization of LLM abilities in speech modality. SALMONN \cite{tang2023salmonn} utilized window-level Q-former \cite{li2023blip2} to glue the pre-trained speech encoder and LLM together. SALMONN exhibits notable emergent inference and reasoning capabilities after being tuned on a large scale audio dataset with various audio/speech/music tasks, employing a multi-stage training curriculum to eventually `activate' the LLM. However, the speech in-context learning capability is not systematically studied.
SLM \cite{wang2023slm} is a successful attempt that enables in-context learning abilities for speech models by bridging the speech representations and the text foundation models. They align these representations by training on supervised ASR, S2TT and speech instruction tuning data that was generated using TTS. The authors evaluate the model's effectiveness in standard speech-related tasks such as ASR and S2TT, while also examining its ability to follow instructions in ASR tasks with contextual biasing. 
Another attempt at building an audio foundation model was Listen Think and Understand (LTU) \cite{gong2023listen}. It was  trained to perform multiple audio (not limited to speech) tasks on a large scale cross-modal question answering dataset. LTU presents emergent audio reasoning and comprehension capabilities, derived from the reasoning abilities of LLMs. 

Besides the integration of speech models and LLMs, there has been exploration for in-context learning in conventional speech model architectures. \cite{lai2023instruction} proposes to apply instruction tuning such as simply ignoring the speech or text manipulation on top of regular speech recognition training, resulting in an ASR model that can execute instruction-guided speech recognition. They show that it is possible to do so without involving a large language model or very large amount of data. Whisper \cite{radford2023robust}, an encoder-decoder model trained on a large-scale weakly-supervised multi-task speech dataset, exhibits preliminary prompt-following capability in its application which is done by feeding free text as prefix to its decoder.  Furthermore, as demonstrated in \cite{hsu2023exploration}, Whisper successfully utilized speech-text paired in-context examples to bias ASR hypotheses, resulting in a decreased Word Error Rate (WER) for various speech dialects.




\begin{figure}[!tbp]
  \centering
  \includegraphics[width=0.45 \textwidth]{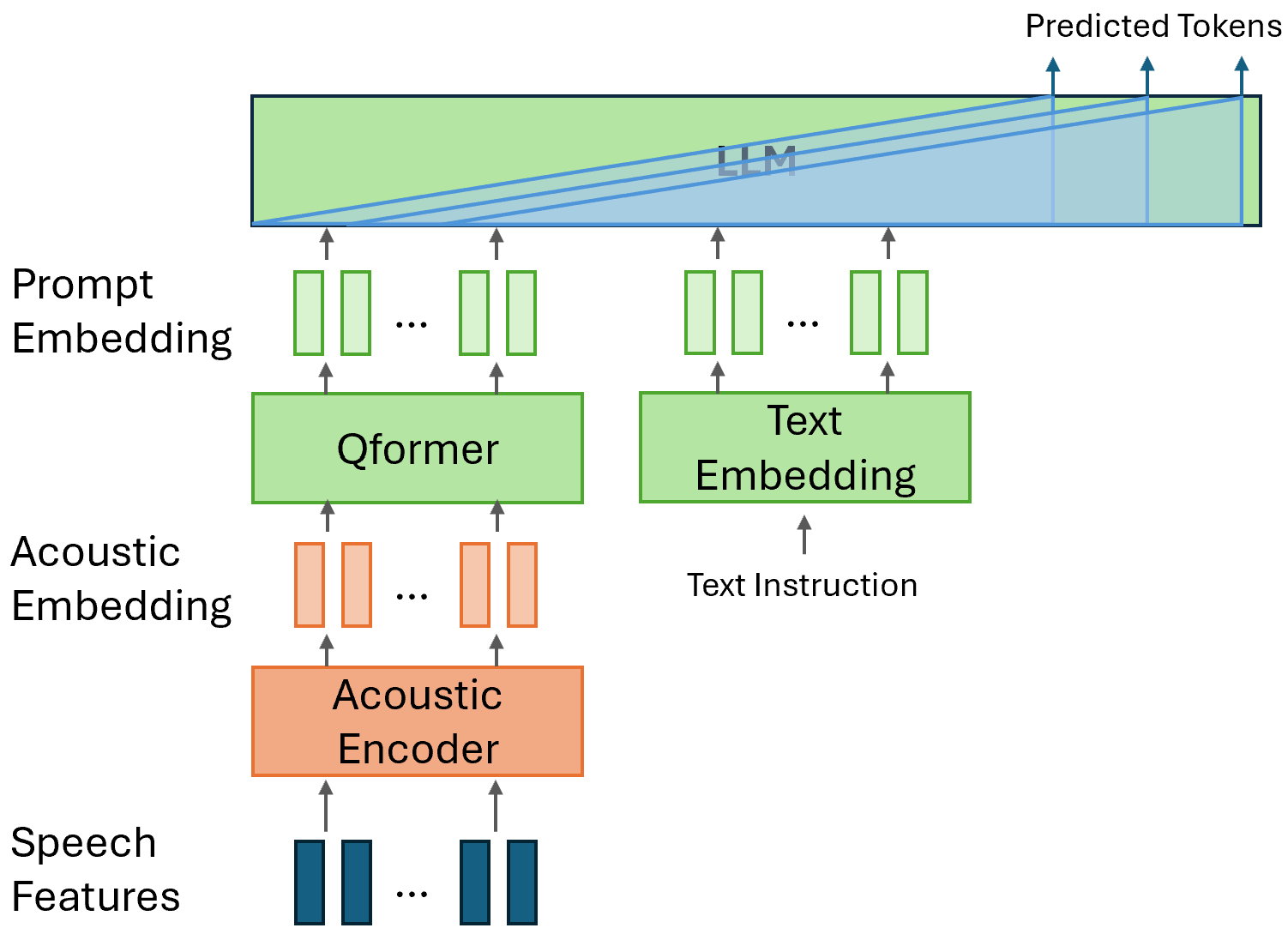}
  \caption{COSMIC architecture}
  \label{fig:modelarch}
  \vspace{-5mm}
\end{figure}

\section{Methodology}

\label{section:methodolgy}
\subsection{Model}
As shown in Figure \ref{fig:modelarch}, there are 3 major blocks in COSMIC model: a pre-trained acoustic encoder, a window-level Qformer \cite{li2023blip2} and an LLM backbone. The speech feature sequence $\mathcal{X}$ will first be fed into an acoustic encoder to obtain an acoustic embedding $\mathcal{M}=\mathtt{Enc} \left(\mathcal{X} \right)$. Window-level Qformer then truncates the acoustic embedding into sub-sequences with a fixed window length $L$, and uses its learnable queries to perform self-attention on each sub-sequences. The output of Qformer can be written as
$$\mathcal{Z}_\text{AcousticPrompt}=[\mathtt{QFormer}(\mathcal{M}_{l})]^{T/L}_{l=1},$$
where $T$ denotes the sequence length and $l$ represents the $l^{th}$ feature chunk. The text instruction will be concatenated with the acoustic prompt embedding. The LLM then takes the prompt embeddings from different modality and decode the prediction in an auto-regressive manner:
$$\mathcal{Y}=\mathtt{LLM} \left(\mathtt{Concat}({Z}_\text{AcousticPrompt}, {Z}_\text{TextPrompt}) \right)$$
In such stack of models, the acoustic encoder maps the raw speech features into higher level acoustic embeddings, then Qformer performs sequence reduction and acoustic and textual modality matching on top of it. During tuning, we enable the Low-Rank Adaptation (LoRA) \cite{hu2021lora} in the LLM to further improve the modality fusion.

\subsection{Speech Instruction-Tuning}
    
  

From our previous exploration, we notice that if we simply perform instruction-tuning for a speech-LLM model on speech-to-text tasks such as ASR or S2TT, the model tends to over-fit to the trained tasks, which means the model wouldn't recognize generalized text instructions other than those seen during the training. In order to leverage the LLMs' strong intrinsic natural language understanding capabilities, we perform the instruction-tuning based on speech comprehension tests, which naturally formulates a one-to-many mapping from the input speech to the \textit{(text instruction, target text response)} tuples. The SQA task uses questions as instructions and allows semantically diverse instructions and target response, since one can raise different questions based on a given speech. Meanwhile, the model is trained to query information from the input speech in SQA, which enhanced the alignment between speech and text modality on top of the ASR task. In Section \ref{section:datasetup}, we describe how we prepare SQA and ASR data for this multi-modal instruction tuning.

\subsection{Speech In-context Learning}
\label{subsection:icl}

After the proposed speech instruction-tuning, COSMIC develops few-shot in-context learning ability that makes it possible to achieve unseen tasks. We enable speech in-context learning during inference stage by positioning context examples as speech and text pairs in the way the COSMIC is trained. 
More formally, with $\{\mathcal{X}_s^0, \cdots, \mathcal{X}_s^{N-1}\}$ as the $N$-shot audio samples, $\{\mathcal{Z}_s^0, \cdots, \mathcal{Z}_s^{N-1}\}$ as embeddings of the corresponding example text outputs and input audio features as $\mathcal{X}_{in}$, the acoustic embedding now becomes 
$$\mathcal{M}=\mathtt{Concat} \left(\text{Enc}(\mathcal{X}_s^0), \cdots, \text{Enc}(\mathcal{X}_s^{N-1}), \text{Enc}(\mathcal{X}_{in}) \right)$$
Finally the prompt fed into the LLM can be written as:
$$\mathtt{Concat} \left(\mathcal{Z}_\text{AcousticPrompt}, \mathcal{Z}_\text{TextPrompt}, \mathcal{Z}_{s}^0, \cdot, \mathcal{Z}_{s}^{N-1} \right)$$
 This type of few-shot prompt is proven to work well for several speech-to-text tasks, such as ASR and S2TT. We discuss how it is evaluated in the sections below.

\section{Data and Evaluation Setup}

\label{section:datasetup}

We use TED-LIUM 3\cite{hernandez2018ted} as our speech data source. This corpus contains 2351 English TED talks, with a total of 452 hours of audio. We send the original transcripts to the OpenAI GPT-3.5 API and ask it to generate questions and the corresponding answers based on each transcript. We expect that longer text context will enable GPT-3.5 to generate more diverse and coherent QAs. Therefore, we use the alignment provided in TED-LIUM 3 to re-segment the audio and get utterances that are about 30 seconds long on average. From this, we obtain 50K audio segments with transcripts and 856K pairs of QAs, which gives us about 17 QA pairs per utterance on average. \footnote{The license of TED-LIUM 3 dataset does not permit distributing the derivatives.}

We evaluate our models on various speech-to-text tasks, which fall into 2 major categories according to whether the model has been trained on those tasks.

\textbf{Trained Tasks:} We test the models on two tasks that they are trained on: ASR and SQA. We use the original TED-LIUM 3 segmentation for ASR and a longer segmentation for SQA, which leads to 1155 and 287 utterances for ASR and SQA respectively. However, one utterance may contain multiple QAs, and there are 1407 QAs in the test set. We measure the semantic similarity between the predicted answers and the GPT-generated answers using BLEU and ROUGE-L metrics. 

\textbf{Unseen Tasks:} The other tasks are unseen during the instruction-tuning and intended to evaluate the models' emergent abilities. We use GPT-3.5 to translate the English transcripts of TED-LIUM 3 test set into Spanish (ES), French (FR) and German (DE) as in-domain S2TT tests, containing 1155 utterances. For a standardized S2TT benchmark, we use the EN$\to$\{ES,FR,DE,ZH(Chinese)\} test set from FLEURS \cite{conneau2023fleurs} with 647 utterances. This task is cross-domain S2TT, since none of our acoustic or LM components is fine-tuned on the FLEURS train set. We also evaluate 1-shot domain adaptation on LibriSpeech \cite{panayotov2015LibriSpeech} test-clean and test-other. We follow the contextual biasing word labels and metrics from \cite{le2021contextualized} and assess how well the model can adhere to the text instructions in the ASR task.

\begin{table}[t]
  \caption{Model Performance on trained tasks (TED-LIUM3)}
  \label{tab:trainedtask}
  \centering
  \begin{tabular}{ cccc }
    \toprule
    \multicolumn{1}{c}{\textbf{Model}} & 
    \multicolumn{1}{c}{\textbf{WER}} &
    \multicolumn{1}{c}{\textbf{BLEU}} &
    \multicolumn{1}{c}{\textbf{ROUGE-L}} \\
    
    \midrule
    Cascaded-7B & 6.67 & 5.14  & 25.92   \\
    COSMIC-ASR-7B & 8.07  & 5.04  & 21.75  \\
    \midrule
    COSMIC-7B & $9.84$ & $33.65$ & $57.41$  \\
    COSMIC-13B & $6.69$ & $35.97$ & $58.95$  \\
    \bottomrule
  \end{tabular}
  
\end{table}

\section{Experiments and Results}
\label{section:experiment}
We perform instruction-tuning on the models with a mixture of ASR and SQA labels. At each step, we sample from the instruction-response set for each utterance with a uniform probability, where the instruction could be either one of the questions from QA pairs or the instruction to transcribe. The input audio feature is 80-dim log mel-filterbank using 25ms window and 10ms hop size. The LLM component of our model is initialized from the LLaMA-2 checkpoints (those without instruct fine-tuning). We port the encoder parameters from Whisper-medium \cite{radford2021learning} to initialize our acoustic encoder. Being trained on large amount of ASR and S2TT data, the Whisper encoder would be a good fit for the tasks evaluated in this work.
There are 2 transformer layers in Qformer with attention dimension set to be 1024, with 1 trainable query vector which is applied to every window of 17 frames.
We set the rank to 2 for LoRA adaptors and train all of our models with 32 V100 GPUs for 250K steps in one stage. Each minibatch contains approximately 64 sequences. We employ the AdamW optimizer and linear learning rate decay with peak learning rate set to $10^{-4}$. We select the checkpoints scoring the best loss on dev set for testing. During instruction-tuning, we freeze the parameters of the acoustic encoder and LLM, leaving only the Qformer and the LoRA adaptors attached to LLM trainable. The trainable components in COSMIC sum up to 27.3 and 28.5 million parameters with 7B and 13B LLM, respectively.

\subsection{Trained Tasks - ASR and SQA}
We assess the model's performance on both ASR and SQA tasks using the TED-LIUM 3 test set, detailed in Table \ref{tab:trainedtask}. For comparison, we showcase the cascaded system's performance, where English audio is decoded with off-the-shelf Whisper-medium, followed by prompting the LLaMA-2 7B for downstream tasks. However, this cascaded system is not entirely apple-to-apple comparable with COSMIC models since the total number parameters are not matched between these 2 setups, counting the decoder of Whisper model. Whisper-medium's Word Error Rate (WER) signifies its ASR performance. In the SQA task, we concatenate the ASR transcript and questions for the out-of-the-box LLaMA-2 models. COSMIC with a 7B LLaMA-2 model is denoted as COSMIC-7B, and COSMIC-13B refers to the COSMIC model with a 13B LLaMA-2 model as the backbone LM. As part of the baseline systems, we train a COSMIC-ASR-7B model for the ASR-only task, outperforming COSMIC-7B in ASR but performing poorly in SQA, as expected. The COSMIC-7B model's gains are attributed to improved comprehension of speech contents through instruction tuning, demonstrating better alignment with the SQA task and achieving higher BLEU and ROUGE-L scores. Switching to a larger 13B LLM as the backbone yields gains in both ASR and SQA tasks, emphasizing the impact of a larger language model.


\begin{table}[t]
  \caption{In-domain EN$\to$X S2TT on TED-LIUM 3 test sets}
  \label{tab:ted_s2tt}
  \setlength{\tabcolsep}{4pt}
  \centering
  \begin{tabular}{ ccccc }
    \toprule
    \multirow{2}{*}{\textbf{Model}} & 
    \multirow{2}{*}{\textbf{\#Example}} & \multicolumn{3}{c}{\textbf{EN$\to$X Target}} \\
    & &
    \multicolumn{1}{c}{\textbf{ES}} &
    \multicolumn{1}{c}{\textbf{FR}} &
    \multicolumn{1}{c}{\textbf{DE}} \\
    \midrule
    Cascaded-7B & - & 19.48 & 17.91 & 12.07\\
    \midrule
    \multirow{2}{*}{COSMIC-ASR-7B} & 0-shot & 10.42 & 11.32 & 7.43 \\
     & 1-shot & 2.80 & 4.53 & 3.58 \\
    \midrule
    \multirow{2}{*}{COSMIC-7B} & 0-shot & $20.63$ & $30.23$ & $20.29$ \\
     & 1-shot & $33.18$ & $30.66$ & $22.16$ \\
    \midrule
    \multirow{2}{*}{COSMIC-13B} & 0-shot & $6.53$ & $9.69$ & $4.66$ \\
     & 1-shot & $\textbf{41.60}$ & $\textbf{36.82}$ & $\textbf{26.83}$ \\
    \bottomrule
  \end{tabular}
\end{table}

\subsection{Unseen Tasks}

    
  

\begin{table}[t]
  \caption{Cross-domain EN$\to$X S2TT on FLEURS test sets}
  \setlength{\tabcolsep}{3pt}
  \label{tab:fleurs_s2tt}
  \centering
  \begin{tabular}{ cccccc }
    \toprule
    \multirow{2}{*}{\textbf{Model}} &  \multirow{2}{*}{\textbf{\#Example}} & \multicolumn{4}{c}{\textbf{EN$\to$X Target}} \\
     & &
    \multicolumn{1}{c}{\textbf{ES}} &
    \multicolumn{1}{c}{\textbf{FR}} &
    \multicolumn{1}{c}{\textbf{DE}} &
    \multicolumn{1}{c}{\textbf{ZH}} \\

    \midrule
    Cascaded-7B & - & 11.37 & 16.63 & 9.45 & 17.22 \\
    \midrule
    \multirow{2}{*}{COSMIC-7B} & 0-shot & $11.09$ & $22.47$ & $13.64$ & $11.00$ \\
     & 1-shot & $15.32$ & $22.83$ & $15.62$ & $18.57$  \\
    \midrule
    \multirow{2}{*}{COSMIC-13B} & 0-shot & $1.65$ & $7.99$ & $3.75$ & $2.74$ \\
     & 1-shot & $17.37$ & $\textbf{27.13}$ & $19.06$ & $\textbf{22.74}$  \\
    \bottomrule
  \end{tabular}
  
\end{table}
\begin{table}[t]
  \caption{One-shot ASR domain adaptation on LibriSpeech}
  \label{tab:libri_adapt}
  \setlength{\tabcolsep}{6pt}
  \centering
  \begin{tabular}{ c l  c l  c l  c l  c}
    \toprule
    \multirow{2}{*}{\textbf{Model}} & 
    \multicolumn{2}{c}{\textbf{test-clean}} &
    \multicolumn{2}{c}{\textbf{test-other}} \\
     & 
    \multicolumn{1}{c}{\textbf{0-shot}} &
    \multicolumn{1}{c}{\textbf{1-shot}} &
    \multicolumn{1}{c}{\textbf{0-shot}} &
    \multicolumn{1}{c}{\textbf{1-shot}} \\
    
    \midrule
    COSMIC-7B & $9.90$ & $5.82$ & $16.05$ & $11.63$ \\
    COSMIC-13B & $7.13$ & $5.53$ & $12.58$ & $10.97$ \\
    \bottomrule
  \end{tabular}
\end{table}

\begin{table}[th]
  \captionsetup{format=myformat}
  \caption{ASR w/ prompt-based contextual biasing}
  \label{tab:libri_cb}
  \setlength{\tabcolsep}{3pt}
  \centering
  \begin{tabular}{ccccccc}
    \toprule
    \multirow{2}{*}{\textbf{Biased}} &
    \multicolumn{3}{c}{\textbf{test-clean}} &
    \multicolumn{3}{c}{\textbf{test-other}}  \\
    & WER & U-WER & B-WER &  WER & U-WER & B-WER \\

    \midrule
    
    \xmark & $9.90$ & $8.40$ & $22.11$ & $16.05$ & $12.71$ & $38.43$ \\
    \cmark & $7.06$ & $6.49$ & $\textbf{11.70}$ & $12.71$ & $11.91$ & $\textbf{19.81}$ \\
    \bottomrule
  \end{tabular}
\end{table}

\subsubsection{EN$\to$X Speech-to-text Translation}
To evaluate the emergent capabilities of COSMIC, we commence with EN$\to$X S2TT. It is important to note that instruction-tuning does not involve non-English text responses. We undertake a comparative analysis of two scenarios: 0-shot versus 1-shot. In the 0-shot scenario, the model is provided with a simple instruction, such as ``\textit{Translate the audio into the \{target language\}}." In the 1-shot scenario, the model is furnished with audio randomly selected from the training set and the corresponding text translation, following the paradigm outlined in Section \ref{subsection:icl}.

Tables \ref{tab:ted_s2tt} and \ref{tab:fleurs_s2tt} present the S2TT results (BLEU scores) for TED-LIUM 3 and FLEURS, respectively. The COSMIC models exhibit commendable 0-shot translation quality, affirming the effectiveness of our instruction-tuning approach in aligning the speech modality with text and cultivating instruction-following capabilities that generalize to \textbf{previously unseen text instructions}. These two factors are pivotal in accomplishing the 0-shot task, and it's noteworthy that this achievement is attained with just 450 hours of English audio data. In Table \ref{tab:ted_s2tt}, we present two baselines: the cascaded system Cascaded-7B and ASR-only model COSMIC-ASR-7B. The cascaded system with LLaMA-2 7B model is outperformed by the COSMIC-7B counterpart with 0-shot inference in most cases, except that in Table \ref{tab:fleurs_s2tt} Cascaded-7B scores higher BLEU scores on some languages, due to the more robust speech transcription capability of Whisper model trained on much larger scale of data. On top of that, the 1-shot example further boosts the translation performance of COSMIC-7B, underscoring the effectiveness of speech in-context learning. Conversely, COSMIC-ASR-7B performs less favorably in both 0-shot and 1-shot S2TT tasks, indicating that SQA task in training is essential for the development of instruction-following and speech in-context learning. In the case of COSMIC-13B, 0-shot translation results are less favorable, potentially due to the need for more data for the 13B model to converge on alignment and instruction following. Notably, there are instances in the COSMIC-13B 0-shot scenario where the model abstains from providing a direct translation and instead generates a query URL pointing to an online translation service API. Nevertheless, with a 1-shot example, COSMIC-13B achieves the highest BLEU score among all languages. 

The overall translation quality deteriorates when inferring on cross-domain datasets, as evident in Table \ref{tab:fleurs_s2tt}. We hypothesize that limited data quantity poses challenges for the model's generalization. However, on both test sets, COSMIC's translation quality experiences a significant boost across the board with just an 1-shot example, demonstrating the effectiveness of COSMIC's speech in-context learning capabilities.

\subsubsection{Cross-domain Adaptation}

Domain adaptation has consistently posed challenges for end-to-end (E2E) ASR models, primarily due to the costly data annotation process inherent in supervised adaptation training methods. However, speech in-context learning offers an avenue for achieving cost-effective, on-the-fly adaptation with just a minimal number of audio-text pairs.
We adopt the speech in-context learning for 1-shot audio domain adaptation. LibriSpeech is selected as the target domain given that COSMIC has not been previously trained on its data. During the inference phase, we randomly select a single utterance from the LibriSpeech training set, along with its corresponding transcription, and provide them as a 1-shot example to COSMIC. This 1-shot example serves as an in-context learning opportunity for COSMIC to adapt to the specific acoustic conditions of the LibriSpeech domain. 
The results of this adaptation process are presented in Table \ref{tab:libri_adapt}. Notably, both of the 7B and 13B COSMIC model exhibits significant reductions in WERs (average 25.8\% relative) across the board with just a single 1-shot example. 

\subsubsection{ASR with Contextual Biasing}
We attempt to showcase the model's instruction-following capability in the contextual biasing scenario. In the set up outlined by \cite{le2021contextualized}, the less common words in LibriSpeech test-\{clean, other\} are identified as the biasing target. During evaluation, WERs are computed independently for the biasing target words and non-target words (termed as B-WER and U-WER respectively). Essentially, B-WER assesses the effectiveness of the biasing process, while U-WER gauges the presence of over-biasing. Additionally, we report the overall WERs for the entire test dataset. For our model, our approach involves conveying contextual information in natural language and feeding it to the model as a text instruction. A typical instruction might read: ``\textit{Transcribe the audio to text. As context, the speaker in the audio mentions $w_1$, $w_2$, ..., and $w_n$}." Here, each word $w_i$ corresponds to one of the target words from the biasing word list. This approach seamlessly integrates the contextual biasing information with the text instruction. We conduct the tests for COSMIC-7B and show the result in Table \ref{tab:libri_cb}. Notably, when utilizing biasing instructions, we observed a significant decrease in B-WER, roughly around 50\%, for both test sets. Furthermore, U-WER drops along with the B-WER. We hypothesize that the LLM is capable of utilizing the given context to clear the confusion and optimize even for un-targeted words. Once again, these results underscore the COSMIC's ability to effectively follow more generalized instructions for controlled text outputs after the proposed instruction-tuning.


\section{Conclusion}
\label{section:conclusion}
In this study, we introduce an data-efficient instruction-tuning approach that leverages ASR and SQA tasks. Our empirical investigations and results substantiate its ability to enhance the emerging skills of instruction-following and in-context speech learning within the speech model. We further illustrate the effectiveness of our instruction-tuned model, named COSMIC, in adhering to instructions and generalizing to unencountered tasks through in-context learning, such as EN$\to$X S2TT, ASR domain adaptation, and ASR with contextual biasing. In  future, we intend to expand our dataset in terms of both its size and task diversity. Additionally, we aim to explore the utility of speech in-context learning for a broader array of applications.


\bibliographystyle{IEEEtran}
\bibliography{main}

\begin{thebibliography}{10}
\providecommand{\url}[1]{#1}
\csname url@samestyle\endcsname
\providecommand{\newblock}{\relax}
\providecommand{\bibinfo}[2]{#2}
\providecommand{\BIBentrySTDinterwordspacing}{\spaceskip=0pt\relax}
\providecommand{\BIBentryALTinterwordstretchfactor}{4}
\providecommand{\BIBentryALTinterwordspacing}{\spaceskip=\fontdimen2\font plus
\BIBentryALTinterwordstretchfactor\fontdimen3\font minus \fontdimen4\font\relax}
\providecommand{\BIBforeignlanguage}[2]{{%
\expandafter\ifx\csname l@#1\endcsname\relax
\typeout{** WARNING: IEEEtran.bst: No hyphenation pattern has been}%
\typeout{** loaded for the language `#1'. Using the pattern for}%
\typeout{** the default language instead.}%
\else
\language=\csname l@#1\endcsname
\fi
#2}}
\providecommand{\BIBdecl}{\relax}
\BIBdecl

\bibitem{brown2020language}
T.~Brown, B.~Mann, N.~Ryder, M.~Subbiah, J.~D. Kaplan, P.~Dhariwal, A.~Neelakantan, P.~Shyam, G.~Sastry, A.~Askell \emph{et~al.}, ``Language models are few-shot learners,'' \emph{Advances in neural information processing systems}, vol.~33, pp. 1877--1901, 2020.

\bibitem{touvron2023llama}
H.~Touvron, L.~Martin, K.~Stone, P.~Albert, A.~Almahairi, Y.~Babaei, N.~Bashlykov, S.~Batra, P.~Bhargava, S.~Bhosale \emph{et~al.}, ``Llama 2: Open foundation and fine-tuned chat models,'' \emph{arXiv preprint arXiv:2307.09288}, 2023.

\bibitem{anil2023palm}
R.~Anil, A.~M. Dai, O.~Firat, M.~Johnson, D.~Lepikhin, A.~Passos, S.~Shakeri, E.~Taropa, P.~Bailey, Z.~Chen \emph{et~al.}, ``Palm 2 technical report,'' \emph{arXiv preprint arXiv:2305.10403}, 2023.

\bibitem{dong2022survey}
Q.~Dong, L.~Li, D.~Dai, C.~Zheng, Z.~Wu, B.~Chang, X.~Sun, J.~Xu, and Z.~Sui, ``A survey for in-context learning,'' \emph{arXiv preprint arXiv:2301.00234}, 2022.

\bibitem{wang2023can}
S.~Wang, C.-H.~H. Yang, J.~Wu, and C.~Zhang, ``Can whisper perform speech-based in-context learning,'' \emph{arXiv preprint arXiv:2309.07081}, 2023.

\bibitem{hsu2023exploration}
M.-H. Hsu, K.-W. Chang, S.-W. Li, and H.-y. Lee, ``An exploration of in-context learning for speech language model,'' \emph{arXiv preprint arXiv:2310.12477}, 2023.

\bibitem{rubenstein2023audiopalm}
P.~K. Rubenstein, C.~Asawaroengchai, D.~D. Nguyen, A.~Bapna, Z.~Borsos, F.~d.~C. Quitry, P.~Chen, D.~E. Badawy, W.~Han, E.~Kharitonov \emph{et~al.}, ``Audiopalm: A large language model that can speak and listen,'' \emph{arXiv preprint arXiv:2306.12925}, 2023.

\bibitem{barrault2023seamlessm4t}
L.~Barrault, Y.-A. Chung, M.~C. Meglioli, D.~Dale, N.~Dong, P.-A. Duquenne, H.~Elsahar, H.~Gong, K.~Heffernan, J.~Hoffman \emph{et~al.}, ``Seamlessm4t-massively multilingual \& multimodal machine translation,'' \emph{arXiv preprint arXiv:2308.11596}, 2023.

\bibitem{baevski2020wav2vec}
A.~Baevski, Y.~Zhou, A.~Mohamed, and M.~Auli, ``wav2vec 2.0: A framework for self-supervised learning of speech representations,'' \emph{Advances in neural information processing systems}, vol.~33, pp. 12\,449--12\,460, 2020.

\bibitem{hsu2021hubert}
W.-N. Hsu, B.~Bolte, Y.-H.~H. Tsai, K.~Lakhotia, R.~Salakhutdinov, and A.~Mohamed, ``Hubert: Self-supervised speech representation learning by masked prediction of hidden units,'' \emph{IEEE/ACM Transactions on Audio, Speech, and Language Processing}, vol.~29, pp. 3451--3460, 2021.

\bibitem{yang21c_interspeech}
S.~wen Yang, P.-H. Chi, Y.-S. Chuang, C.-I.~J. Lai, K.~Lakhotia, Y.~Y. Lin, A.~T. Liu, J.~Shi, X.~Chang, G.-T. Lin, T.-H. Huang, W.-C. Tseng, K.~tik Lee, D.-R. Liu, Z.~Huang, S.~Dong, S.-W. Li, S.~Watanabe, A.~Mohamed, and H.~yi~Lee, ``{SUPERB: Speech Processing Universal PERformance Benchmark},'' in \emph{Proc. Interspeech 2021}, 2021, pp. 1194--1198.

\bibitem{fathullah2023prompting}
Y.~Fathullah, C.~Wu, E.~Lakomkin, J.~Jia, Y.~Shangguan, K.~Li, J.~Guo, W.~Xiong, J.~Mahadeokar, O.~Kalinli \emph{et~al.}, ``Prompting large language models with speech recognition abilities,'' \emph{arXiv preprint arXiv:2307.11795}, 2023.

\bibitem{lakomkin2023end}
E.~Lakomkin, C.~Wu, Y.~Fathullah, O.~Kalinli, M.~L. Seltzer, and C.~Fuegen, ``End-to-end speech recognition contextualization with large language models,'' \emph{arXiv preprint arXiv:2309.10917}, 2023.

\bibitem{wang2023slm}
M.~Wang, W.~Han, I.~Shafran, Z.~Wu, C.-C. Chiu, Y.~Cao, Y.~Wang, N.~Chen, Y.~Zhang, H.~Soltau \emph{et~al.}, ``Slm: Bridge the thin gap between speech and text foundation models,'' \emph{arXiv preprint arXiv:2310.00230}, 2023.

\bibitem{yu2023connecting}
W.~Yu, C.~Tang, G.~Sun, X.~Chen, T.~Tan, W.~Li, L.~Lu, Z.~Ma, and C.~Zhang, ``Connecting speech encoder and large language model for asr,'' \emph{arXiv preprint arXiv:2309.13963}, 2023.

\bibitem{wu2023decoder}
J.~Wu, Y.~Gaur, Z.~Chen, L.~Zhou, Y.~Zhu, T.~Wang, J.~Li, S.~Liu, B.~Ren, L.~Liu \emph{et~al.}, ``On decoder-only architecture for speech-to-text and large language model integration,'' \emph{arXiv preprint arXiv:2307.03917}, 2023.

\bibitem{liu2023visual}
H.~Liu, C.~Li, Q.~Wu, and Y.~J. Lee, ``Visual instruction tuning,'' \emph{arXiv preprint arXiv:2304.08485}, 2023.

\bibitem{zhang2023speechgpt}
D.~Zhang, S.~Li, X.~Zhang, J.~Zhan, P.~Wang, Y.~Zhou, and X.~Qiu, ``Speechgpt: Empowering large language models with intrinsic cross-modal conversational abilities,'' \emph{arXiv preprint arXiv:2305.11000}, 2023.

\bibitem{tang2023salmonn}
C.~Tang, W.~Yu, G.~Sun, X.~Chen, T.~Tan, W.~Li, L.~Lu, Z.~Ma, and C.~Zhang, ``Salmonn: Towards generic hearing abilities for large language models,'' \emph{arXiv preprint arXiv:2310.13289}, 2023.

\bibitem{li2023blip2}
J.~Li, D.~Li, S.~Savarese, and S.~Hoi, ``Blip-2: Bootstrapping language-image pre-training with frozen image encoders and large language models,'' 2023.

\bibitem{gong2023listen}
Y.~Gong, H.~Luo, A.~H. Liu, L.~Karlinsky, and J.~Glass, ``Listen, think, and understand,'' \emph{arXiv preprint arXiv:2305.10790}, 2023.

\bibitem{lai2023instruction}
C.-I.~J. Lai, Z.~Lu, L.~Cao, and R.~Pang, ``Instruction-following speech recognition,'' \emph{arXiv preprint arXiv:2309.09843}, 2023.

\bibitem{radford2023robust}
A.~Radford, J.~W. Kim, T.~Xu, G.~Brockman, C.~McLeavey, and I.~Sutskever, ``Robust speech recognition via large-scale weak supervision,'' in \emph{International Conference on Machine Learning}.\hskip 1em plus 0.5em minus 0.4em\relax PMLR, 2023, pp. 28\,492--28\,518.

\bibitem{hu2021lora}
E.~J. Hu, Y.~Shen, P.~Wallis, Z.~Allen-Zhu, Y.~Li, S.~Wang, L.~Wang, and W.~Chen, ``Lora: Low-rank adaptation of large language models,'' \emph{arXiv preprint arXiv:2106.09685}, 2021.

\bibitem{hernandez2018ted}
F.~Hernandez, V.~Nguyen, S.~Ghannay, N.~Tomashenko, and Y.~Esteve, ``Ted-lium 3: Twice as much data and corpus repartition for experiments on speaker adaptation,'' in \emph{Speech and Computer: 20th International Conference, SPECOM 2018, Leipzig, Germany, September 18--22, 2018, Proceedings 20}.\hskip 1em plus 0.5em minus 0.4em\relax Springer, 2018, pp. 198--208.

\bibitem{conneau2023fleurs}
A.~Conneau, M.~Ma, S.~Khanuja, Y.~Zhang, V.~Axelrod, S.~Dalmia, J.~Riesa, C.~Rivera, and A.~Bapna, ``Fleurs: Few-shot learning evaluation of universal representations of speech,'' in \emph{2022 IEEE Spoken Language Technology Workshop (SLT)}.\hskip 1em plus 0.5em minus 0.4em\relax IEEE, 2023, pp. 798--805.

\bibitem{panayotov2015LibriSpeech}
V.~Panayotov, G.~Chen, D.~Povey, and S.~Khudanpur, ``Librispeech: an asr corpus based on public domain audio books,'' in \emph{2015 IEEE international conference on acoustics, speech and signal processing (ICASSP)}.\hskip 1em plus 0.5em minus 0.4em\relax IEEE, 2015, pp. 5206--5210.

\bibitem{le2021contextualized}
D.~Le, M.~Jain, G.~Keren, S.~Kim, Y.~Shi, J.~Mahadeokar, J.~Chan, Y.~Shangguan, C.~Fuegen, O.~Kalinli \emph{et~al.}, ``Contextualized streaming end-to-end speech recognition with trie-based deep biasing and shallow fusion,'' \emph{arXiv preprint arXiv:2104.02194}, 2021.

\bibitem{radford2021learning}
A.~Radford, J.~W. Kim, C.~Hallacy, A.~Ramesh, G.~Goh, S.~Agarwal, G.~Sastry, A.~Askell, P.~Mishkin, J.~Clark \emph{et~al.}, ``Learning transferable visual models from natural language supervision,'' in \emph{International conference on machine learning}.\hskip 1em plus 0.5em minus 0.4em\relax PMLR, 2021, pp. 8748--8763.

\end{thebibliography}
\end{document}